\documentclass[10pt,twocolumn,letterpaper]{article}

\usepackage{cvpr}
\usepackage{times}
\usepackage{epsfig}
\usepackage{graphicx}
\graphicspath{{./images/}}
\usepackage{tabulary}
\usepackage{booktabs}
\usepackage{multicol}
\usepackage{multirow}
\usepackage{amsmath}
\usepackage{amssymb}
\usepackage{gensymb}
\usepackage{comment}
\usepackage{xcolor}
\usepackage{appendix}

\usepackage[numbers]{natbib}

\usepackage[pagebackref=true,breaklinks=true,letterpaper=true,colorlinks,bookmarks=false]{hyperref}

\cvprfinalcopy 

\def\cvprPaperID{7262} 

\ifcvprfinal\fi
\begin{document}

\title{End-to-End Model-Free Reinforcement Learning\\for Urban Driving using Implicit Affordances}


\author{%
  Marin Toromanoff\textsuperscript{1,2,3}, Emilie Wirbel\textsuperscript{2,3}, Fabien Moutarde\textsuperscript{1} \\
  \small{\textsuperscript{1}Center for Robotics, MINES ParisTech, PSL\ \ \ \ \ \ \ \ \ \ \ \  \textsuperscript{2}Valeo Driving Assistance Research\ \ \ \ \ \ \ \ \ \ 
  \textsuperscript{3}Valeo.ai} \\
  \small{\textsuperscript{1}\texttt{name.surname@mines-paristech.fr}\ \ \ \ \ \ \ \ \ \ \ \ 
  \textsuperscript{2, 3}\texttt{name.surname@valeo.com}}
}

\maketitle

\begin{abstract}

Reinforcement Learning (RL) aims at learning an optimal behavior policy from its own experiments and not rule-based control methods. However, there is no RL algorithm yet capable of handling a task as difficult as urban driving. 
We present a novel technique, coined \textit{implicit affordances}, to effectively leverage RL for urban driving thus including lane keeping, pedestrians and vehicles avoidance, and traffic light detection. To our knowledge we are the first to present a successful RL agent handling such a complex task especially regarding the traffic light detection. Furthermore, we have demonstrated 
the effectiveness of our method by winning the Camera Only track of the CARLA challenge.
\end{abstract}

\section{Introduction}

Urban driving is probably one of the hardest situations to solve for autonomous cars, particularly regarding the interaction on intersections with traffic lights, pedestrians crossing and cars going on different possible lanes. Solving this task is still an open problem and it seems complicated to handle such difficult and highly variable situations with classic rules-based approach. This is why a significant part of the state of the art in autonomous driving \cite{Liang, Codevilla, CILRS} focuses on end-to-end systems, i.e.\ learning driving policy from data without relying on hand-crafted rules.

Imitation learning (IL) \cite{Pomerleau1989a} aims to reproduce the behavior of an expert (a human driver for autonomous driving) by learning to mimic the control the human driver applied in the same situation. This leverages the massive amount of data annotated with human driving that most of automotive manufacturer and supplier can obtain relatively easily. On the other side, as the human driver is always in an almost perfect situation, IL algorithms suffer from a distribution mismatch, i.e.\ the algorithm will never encounter failing cases and thus will not react appropriately in those conditions. Techniques to augment the database with such failing cases do exist but they are currently mostly limited to lane keeping and lateral control \cite{Bojarski2016d, toromanoff2018end}.  

Deep Reinforcement Learning (DRL) on the other side lets the algorithm learn by itself by providing a reward signal at each action taken by the agent and thus does not suffer from distribution mismatch. This reward can be sparse and not describing exactly what the agent should have done but just how good the action taken is locally. The final goal of the agent is to maximize the sum of accumulated rewards and thus the agent needs to think about sequence of actions rather than instantaneous ones. One of the major drawbacks of DRL is that it can need a magnitude larger amount of data than supervised learning to converge, which can lead to difficulties when training large networks with many parameters. Moreover many RL algorithms rely on a replay buffer \cite{Lillicrap2015, mnih2015human, Rainbow} allowing to learn from past experiments but such buffers can limit the size of the input used (e.g. the size of the image). That is why neural networks and image size in DRL are usually tiny compared to the ones used in supervised learning. Thus they may not be expressive enough to solve such complicated tasks as urban driving. Therefore current DRL approaches to autonomous driving are applied to simpler cases, e.g. only steering control for lane keeping \cite{Kendall} or going as fast as possible in racing games \cite{Mnih, jaritz2018end}. 
Another drawback of DRL, shared with IL, is that the algorithm appears as a black box from which it is difficult to understand how the decision was taken.

A promising way to solve both the data efficiency (particularly for DRL) and the black box problem is to use privileged information as auxiliary losses also coined affordances in some recent papers \cite{Chen, Sauer}. The idea is to train a network to predict high level information such as semantic segmentation maps, distance to center of the lane, traffic light state etc... This prediction can then be used in several ways, either by a classic controller as in Sauer et al. \cite{Sauer}, either as auxiliary loss helping to find better features to the main imitative task loss as in Mehta et al. \cite{Mehta} or also in a model-based RL approach as in the really recent work of Pan et al. \cite{Pan2019a} while also providing some interpretable feedback on how the decision was taken.

In this work, we will present our RL approach for the case of end-to-end urban driving from vision, including lane keeping, traffic light detection, pedestrian and vehicle avoidance, and handling intersection with incoming traffic. To achieve this we introduce a new technique that we coin \textit{implicit affordances}. The idea is to split the training in two phases: first an encoder backbone (Resnet-18 \cite{Resnet}) is trained to predict affordances such as traffic light state or distance to center of the lane. Then the output features of this encoder is used as the RL state instead of the raw images. Therefore the RL signal is only used to train the last part of the network. Moreover the features are used directly in the replay memory rather than the raw images, which corresponds to approximately 20 times less memory needed. We showed our method performance by winning the ``Camera Only" track in the CARLA Autonomous Driving Challenge \cite{carlaChallenge}. To our knowledge we are the first to show a successful RL agent on urban driving, particularly with traffic lights handling.

We summarize our main contributions below:

\begin{itemize}
    \item The first RL agent successfully driving from vision in urban environment including intersection management and traffic lights detection.
    \item Introducing a new technique coined \textit{implicit affordances} allowing training of replay memory based RL with much larger network and input size than most of network used in previous RL works.
    \item Extensive parameters and ablation studies of implicit affordances and reward shaping.
    \item Showcase of the capability or our method by winning the ``Camera Only" track in the CARLA Autonomous Driving Challenge.
\end{itemize}

\section{Related Work}

\subsection{End-to-End Autonomous Driving with RL}

As RL relies on trial and error, most of RL works applied to autonomous cars are conducted in simulation both for safety reasons and data efficiency. One of the most used simulator is TORCS \cite{Wymann2015} as it is an open-source and simple to use racing game. Researchers used it to test their new actor-critic algorithm to control a car with discrete actions in Mnih et al. \cite{Mnih} and with continuous actions in Lillicrap et al. \cite{Lillicrap2015}. But as TORCS is a racing game, the goal of those works is to reach the end of the track as fast as possible and thus does not handle intersections nor traffic lights.

Recently, many papers used the new CARLA \cite{Dosovitskiy} simulator as an open-source urban simulation including pedestrians, intersections and traffic lights. In the original CARLA paper \cite{Dosovitskiy}, the researchers released a driving benchmark along with one Imitation learning and one RL baseline. The RL baseline was using the A3C algorithm with discrete actions \cite{Mnih} and its results were far behind the imitation baseline. Lang et al \cite{Liang} used RL with DDPG \cite{Lillicrap2015} and continuous actions to fine-tune an imitation agent. But they rely mostly on imitation learning and do not explicitly explain how much improvement comes from the RL fine-tuning. Moreover they also do not handle traffic lights. 

Finally, there are still only few RL methods applied in a real car. The first one was Learning to Drive in a Day \cite{Kendall} in which an agent is trained directly on the real car for steering. A really recent work \cite{Zej} also integrates RL on a real car and compares different ways of transferring knowledge learned in CARLA in the real world. Even if their studies are really interesting, their results are preliminary and applied only on few specific real-world scenarios. Both of these works only handle steering angle for lane keeping and a large gap has to be crossed before reaching throttle and steering control simultaneously in urban environment on a real car with RL.

\subsection{Auxiliary Tasks and Learning Affordances}

The UNREAL agent \cite{Jaderberg} is one of the first articles to study the impact of auxiliary tasks for DRL. They showed that adding losses such as predicting incoming reward could improve data efficiency and final performance on both Atari games and labyrinth exploration.

Chen et al. \cite{Chen} introduce affordance prediction for autonomous driving: a neural network is trained to predict high level information such as distance to the right, center and left part of the lane or distance to the preceding car. Then they used those affordances as input to a rule-based controller and reached good performance on the racing simulator TORCS. Sauer et al. upgraded this in their Conditionnal Affordance Learning \cite{Sauer} paper to handle more complicated scenarios such as urban driving. In order to achieve that they also predict information specific to urban driving such as the maximum allowed speed and the incoming traffic light state. As Chen et al. they finally used those information in a rule-based controller and showed their performance in the CARLA benchmark \cite{Dosovitskiy} for urban driving. Both of those works do not include any RL and rely on rule-based controller.
Just after, Mehta et al. \cite{Mehta} used affordances as auxiliary tasks to their imitation learning agent and showed it was improving both data efficiency and final performance. But they do not handle traffic lights and rely purely on imitation.

Finally, there are two really recent articles closely related to ours. The first one by Gordon et al \cite{Gordon} introduced SplitNet on which they explicitly decompose the learning scheme in finding features from perception task and use these features as input to their model-free RL agent. But their scheme is applied to a completely different task, robot navigation and scene exploration.
The second one by Pan et al. \cite{Pan2019a} train a network to predict high-level information such as probability of collision or being off-road in the near futures from a sequence of observations and actions. They use this network in a model-based RL scheme by evaluating different trajectories to finally apply the generated trajectory giving the lowest cost. However, they use a model-based approach and do not handle traffic light signal.

\section{The CARLA Challenge}

\begin{figure}[t]
    \centering
    \begin{tabular}{cc}
         \includegraphics[width=108px]{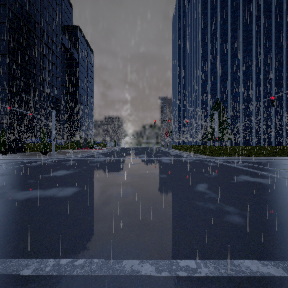} & \includegraphics[width=108px]{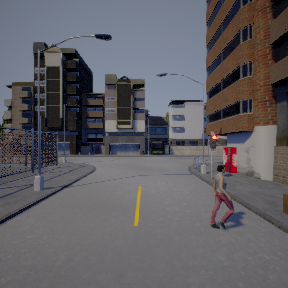} \\
    \end{tabular}
    \caption{Sample of traffic light image (left is US, right is EU).}
    \label{fig:traffic_light}
\end{figure}

The CARLA Challenge \cite{carlaChallenge} is an open competition for autonomous driving relying on the CARLA simulator. This competition addresses specifically the problem of urban driving. The goal is to drive in unseen maps from sensors to control, ensuring lane keeping, handling intersections with high level navigation orders (Right, Left, Straight), handling lane changes, pedestrians and other vehicles avoidance and finally handling traffic lights US and EU at the same time (traffic lights are positioned differently in Europe and in US, see Figure~\ref{fig:traffic_light}). This is much more challenging than the original CARLA benchmark \cite{Dosovitskiy}. The CARLA Challenge consists in 4 different tracks with the only difference being the sensors available, from cameras only to a full stack perception. We will only handle the ``Camera Only" track there, in fact we even used only a single frontal camera for all this work.

\section{Method}

In this section we describe our general approach (RL setup, reward shaping and network architecture). In the next section, we describe what adaptations are needed to make this approach usable in an autonomous driving context.

\subsection{RL Setup: Rainbow-IQN Ape-X}

There are two main families of model-free RL: value-based and policy-based methods.
We choose to use value-based RL as it is the current state-of-the-art on Atari \cite{Rainbow} and is known to be more data efficient than policy-based method. However, it has the drawback of handling only discrete actions.
Making a comparison between value-based RL and policy-based RL (or actor-critic RL which is a sort of combination of both) for Urban driving is out of the scope of this paper but would definitely be interesting for future work.
We started with 
our open-source
\footnote{https://github.com/valeoai/rainbow-iqn-apex}{ implementation} 
of Rainbow-IQN Ape-X \cite{Rainbow, Dabney2018, Horgan} (for Atari originally) taken 
from our previous work 
\cite{Toromanoff}. We removed the dueling network \cite{Dueling} from Rainbow as we found it was leading to same performance while using much more parameters.
The distributed version of Rainbow-IQN was mandatory for our usage: CARLA is too slow for RL and cannot generate enough data if only one instance is used. Moreover this allowed us to train on multiple maps of CARLA at the same time, generating more variability in the training data, better exploration and providing an easy way to handle both US and EU traffic lights (some town used in training were US while others were EU).

\subsection{Reward Shaping}

\begin{figure}[t]
    \centering
    \includegraphics[width=0.5\textwidth]{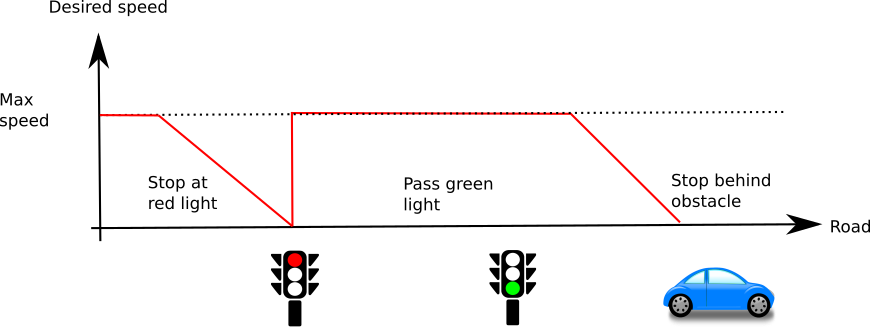}
    \caption{Desired speed according to environment. The desired speed adapts in function of the situation, getting lower when arriving close to a red light, going back to maximum speed when traffic light goes to green and again getting lower when arriving behind an obstacle. The speed reward is maximum when the vehicle speed is equal to the desired speed.}
    \label{fig:reward_speed}
\end{figure}

The reward used for the training relies mostly on the waypoint API present in the latest version of CARLA (CARLA 0.9.X). This API allows to get continuous waypoints position and orientation of all lanes in the current town. This is fundamental to decide what path the agent has to follow. Moreover, this API provides the different possibilities at each intersection. At the beginning of an episode, the agent is initialized on a random waypoint on the city, then the optimal trajectory the agent should follow can be computed using the waypoint API. When arriving at an intersection, we choose randomly a possible maneuvre (Left, Straight or Right) and the corresponding order is given to the agent. The reward relies on three main components: \textit{desired speed}, \textit{desired position} and \textit{desired rotation}. 

The \textit{desired speed} reward is maximum (and equal to 1) when the agent is at the desired speed, and linearly goes down to 0 if the agent speed is lower or higher. The desired speed, illustrated on Figure \ref{fig:reward_speed}, is adapting to the situation: when the agent arrives near a red traffic light, the desired speed goes linearly to 0 (the closest the agent is from the traffic light), and goes back to maximum allowed speed when it turns green. The same principle is used when arriving behind an obstacle, pedestrian, bicycle or vehicle. The desired speed is set to a constant \textit{maximum speed} (here 40km/h) on all other situations.

The second part of the reward, the \textit{desired position}, is inversely proportional to the distance from the middle of the lane (we compute this distance using the waypoints mentioned above). This reward is maximum equal to 0 when agent is exactly in the middle of the lane and goes to -1 when reaching a maximum distance from lane $D_\text{max}$. When the agent is further than $D_\text{max}$, the episode terminates. For all our experiments, $D_\text{max}$ was set to 2 meters: this is the distance from the middle of the lane to the border. Other termination conditions are colliding with anything, running a red light and being stuck for no reason (i.e. not behind an obstacle nor stopped at a red traffic light). For all those termination conditions, the agent receives a reward of -1.

\begin{figure}[t]
    \centering
    \includegraphics[width=0.65\columnwidth]{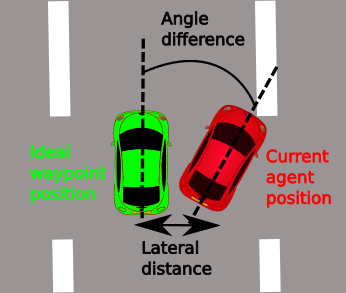}
    \caption{Lateral distance and angle difference for lateral and angle reward computation. The difference is measured between the ideal waypoint (in green) and the current agent position (in red).}
    \label{fig:reward_lat}
\end{figure}

With only the two previous reward components, we observed the trained agents were not going straight as oscillations near the center of lane were giving almost the same amount of reward as going straight. That is why we added our third reward component, \textit{desired rotation}. This reward is inversely proportional to the difference in angle between the agent and the orientation of the nearest waypoint from the optimal trajectory (see Figure~\ref{fig:reward_lat} for details). Ablation studies on the reward shaping can be found at section~\ref{ablation_RL}.


\subsection{Network Architecture}

\begin{figure*}[ht]
    \centering
    \includegraphics[width=0.8\textwidth]{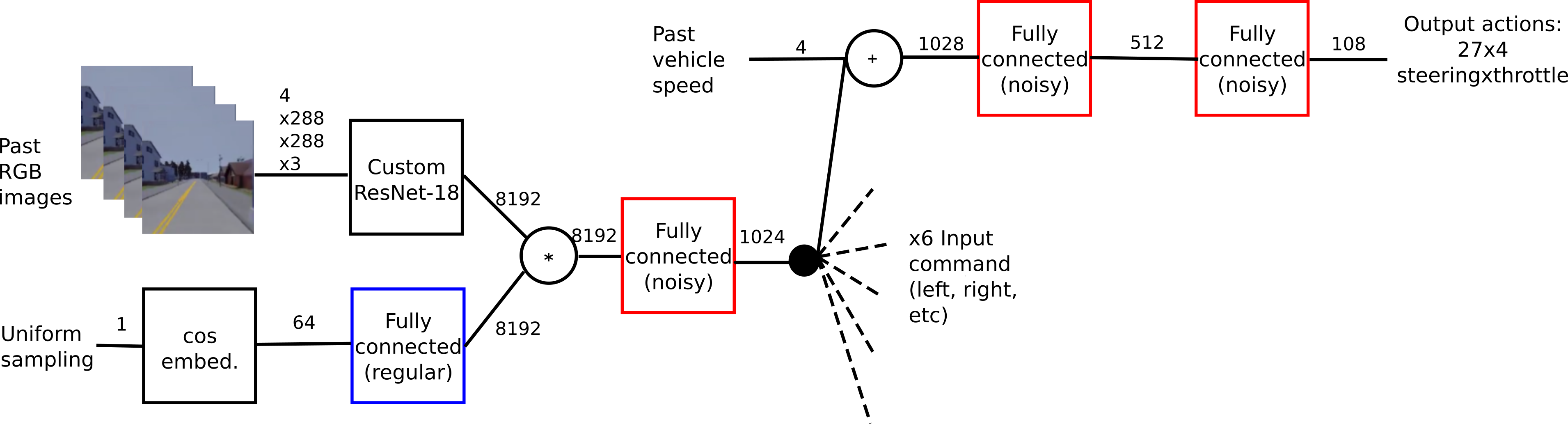}
    \caption{Network architecture. A Resnet-18 \cite{Resnet} encoder is used in a conditional network \cite{Codevilla} with a Rainbow-IQN \cite{Toromanoff} RL training (hence the IQN network \cite{Dabney2018} and noisy fully connected layers \cite{NoisyNetwork})}
    \label{fig:network_archi}
\end{figure*}

Most of networks used in model-free RL with images as input train a particularly small network \cite{Dosovitskiy, Rainbow} compared to networks used commonly in supervised learning \cite{VGG, Resnet}. One of the larger networks used for model-free RL for Atari is the \textit{large architecture} from IMPALA \cite{IMPALA} which consists of 15 convolutional layers and 1.6 million parameter: as comparison our architecture has 18 convolutional layers and 30M parameters. Moreover IMPALA used more than 1B frames when we used only 20M. The most common architecture (e.g. \cite{Mnih, Dabney2018}) is the one introduced in the original DQN paper \cite{mnih2015human}, taking a $84 \times 84$ grayscale image as input.
Our first observation is that traffic light state (particularly for US traffic lights which are farther) can not be seen on so small images. 
Therefore a larger input size has been chosen (around 40 times larger): $4\times288\times288\times3$ by concatenating 4 consecutive frames as a simple and standard \cite{mnih2015human, Dosovitskiy} way to add some temporality in the input. We choose this size as it was the smallest one we tested on which we still had a good accuracy on traffic light detection (using a conventional supervised training).
We choose to use Resnet-18 \cite{Resnet} as a relatively small network (compared to the one used in supervised training) to ensure a small inference time. Indeed RL needs a lot of data to converge so each step must be as fast as possible to reduce the overall training time. However, even if Resnet-18 is among the smallest networks used for supervised learning, it contains around 140 times more weights in its convolutional layers than DQN \cite{mnih2015human}. Moreover Resnet-18 incorporates most of state-of-the art advances in supervised learning such as residual connections and batchnorm \cite{Batchnorm}. Finally, we use a conditional network as in Codevilla et al. \cite{Codevilla} to handle 6 different maneuvers: follow lane, left/right/straight, change lane left/right. The full network architecture is described in Figure~\ref{fig:network_archi}.

\section{Challenges and Solutions to apply RL to Complex Autonomous Driving Tasks}

In this section, we present our suggestions to solve the issues arising when using a large network with RL and how to handle discrete actions.

\subsection{Training RL with high complexity input size: Implicit Affordances}
\label{frozen_encoder}


\paragraph{How to train a larger network with larger images for RL?}
Using a larger network and input size raises two major issues. The first one is that such a network is much longer and harder to train. Indeed it is well known that training a DRL agent is data consuming even with tiny networks. The second issue is the replay memory. One of the major advantages of value-based RL \cite{mnih2015human, Rainbow} over policy-based methods is to be off-policy, meaning the data used for learning can come from another policy.
However storing image 35 times bigger raises issues for storing as many transitions (usually 1M transitions are stored which correspond to 6GB for $84 \times 84$ images and thus would be 210GB for $288 \times 288 \times 3$ images which is unpractical).

Our main idea is to pre-train the convolutional encoder part of the network to predict some high-level information and then freeze it while training the RL. The intuition is that the RL signal is too weak to train the whole network but can be used to train only the fully connected part. Moreover this solves the replay memory issue as we can now store features directly in the replay memory and not the raw images. We coin this scheme as \textit{implicit affordances} because the RL agent does not use the explicit predictions but has only access to the implicit features (i.e the features from which our initial supervised network predicts the explicit affordances).

\paragraph{Which high level semantic information/affordances to predict?}

The most simple idea to pre-train our encoder would be to use an auto-encoder \cite{VAE}, i.e. trying to compress the images by trying to predict back the full image from a smaller feature space. This was used in the work Learning to Drive in a Day \cite{Kendall} and allowed for faster training on their real car. We argue this would not work for our harder use-case particularly regarding the traffic light detection. Indeed, traffic light states represent only a few pixels in the image (red or green) but those pixels are the most relevant for the driving behavior. 

\begin{figure}[t]
    \centering
    \includegraphics[width=0.5\textwidth]{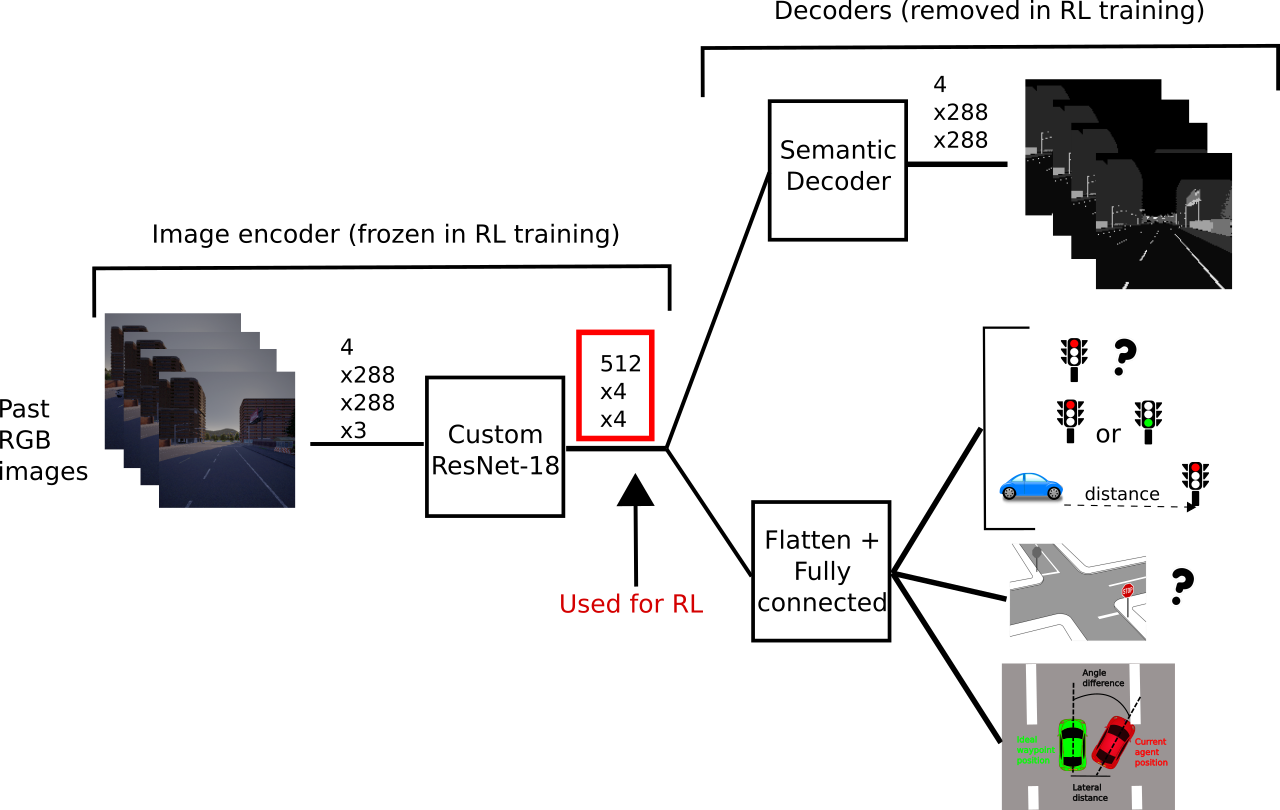}
    \caption{Decoder and losses used to train the encoder: semantic segmentation, traffic light (presence, state, distance), intersection presence, lane position (distance and rotation)}
    \label{fig:encoder_losses}
\end{figure}

To ensure that there is relevant signal in the features used as RL state, we choose to rely on high level semantic information available in CARLA.
We use 2 main losses for our supervised phase: traffic light state (binary classification) and semantic segmentation. Indeed all relevant information but traffic light state is contained in our semantic segmentation. We use 6 classes for the semantic mask: moving obstacles, traffic lights, road markers, road, sidewalk and background. We also predict some other affordances to help the supervised training such as the distance to the incoming traffic light, if we are in an intersection or not, the distance from the middle of the lane and the relative rotation to the road. The two last estimations are coming from our viewpoint augmentation (without it the autopilot is always perfectly in the middle of the lane with no rotation). Our supervised training with all our losses is represented in Figure~\ref{fig:encoder_losses}. Ablation studies to estimate the impact of these affordance estimations are presented on section~\ref{ablation_supervised}.

\paragraph{Viewpoints Augmentation}

\begin{figure}[t]
    \centering
    \includegraphics[width=0.9\columnwidth]{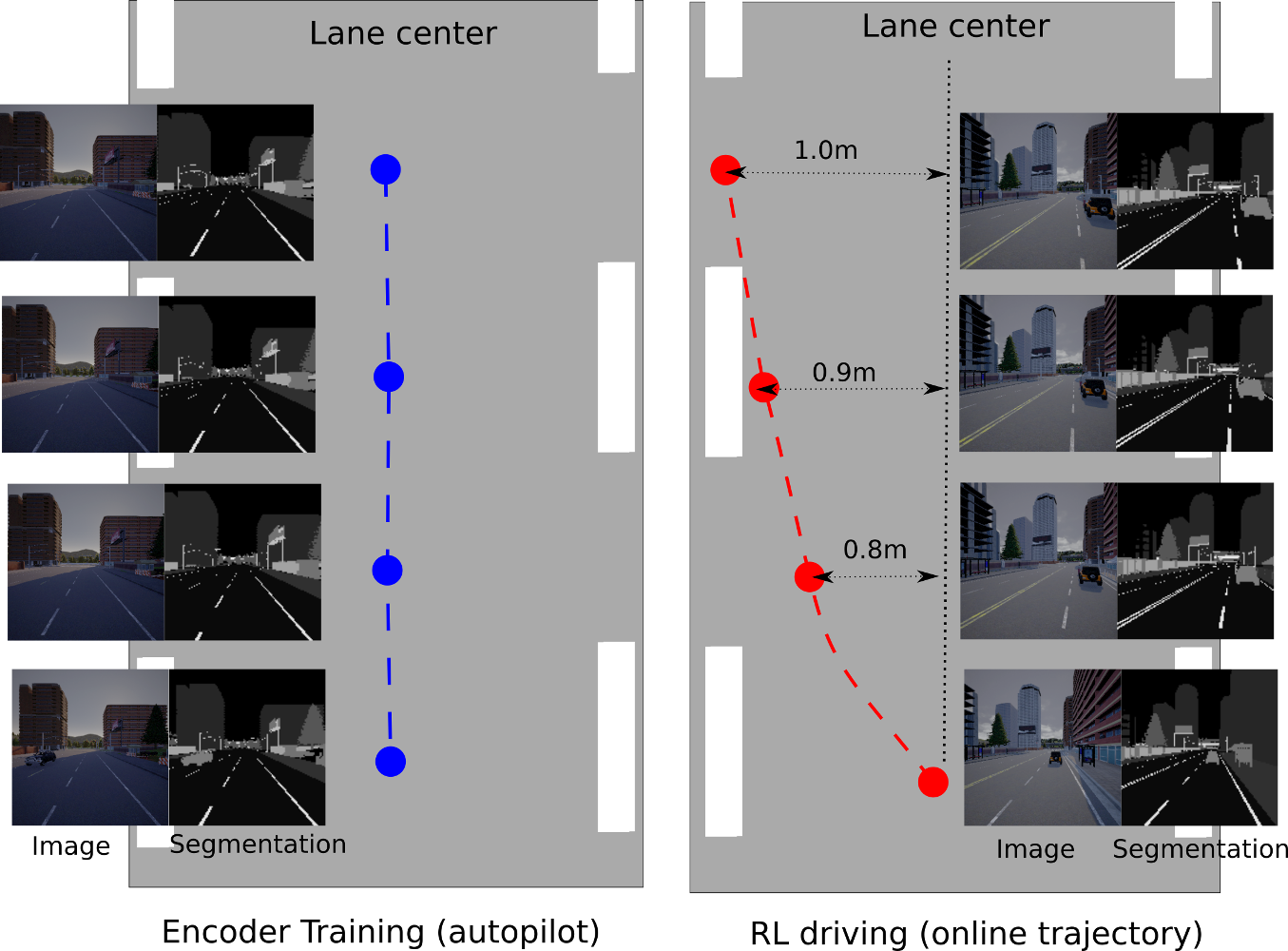}
    \caption{Why data augmentation is needed for training the encoder: RL agents trajectories (right) might deviate from the lane center, which leads to semantic segmentation with much more varied lane marking positions than what can be encountered if training only from autopilot data (left).}
    \label{fig:data_augm}
\end{figure}

The data for the supervised phase is collected while driving with an existing autopilot in the CARLA simulator. However this autopilot always stays in the middle of the lane, so the pre-trained encoder which is frozen does not generalize well during the RL training, particularly when the agent starts to deviate from the middle of the lane: with an encoder trained on data collected only from autopilot driving, an RL agent performance is poor. 
This is the exact same idea as for IL with the distribution mismatch and the intuition behind it is explained on Figure~\ref{fig:data_augm}. To solve this, we suggest to add viewpoints augmentation by moving the camera around the autopilot. With this augmentation the encoder performance is much better while the RL agent drives and explores and we found this was mandatory to obtain good performance during the RL training phase.

In summary, training a large encoder with well selected supervised tasks and using the resulting feature maps, the implicit affordances, as an input state for RL training, addresses the problem of input, network and replay memory size. One should take care to properly augment the data during the supervised training phase, to make sure the encoding is adapted during the exploration of the RL training.

\subsection{Handling Discrete Actions}

As aforementioned, standard value-based RL algorithms such  as DQN \cite{mnih2015human}, Rainbow \cite{Rainbow} and Rainbow-IQN \cite{Toromanoff} imply to use discrete actions. Preliminary experiment with few discrete actions (only 5 for steering) resulted in agents oscillating and failing to stay in lane.
Better results can be obtained by using more steering actions such as 9 or 27 different steering values. Throttle is less of an issue: 3 different values for throttle are used, plus one for brake. This leads to a total of 36 ($9 \times 4$) or 108 ($27 \times 4$) actions for our experiments. We also try to predict the derivative of steering angle: the prediction of network is used to update the previous steering (which is given as input) instead of using directly the prediction as current steering. The impact of these choices is studied in section~\ref{ablation_RL}.

To reach more fine-grained discrete actions, we strongly suggest to use a bagging of multiple predictions and average them. 
To do so, we can simply use consecutive snapshots of the same training, which avoids having to train again and is free to have. This trick is consistently improving behavior, reducing oscillations by a large margin and obtaining better final performance. Furthermore, as the encoder is frozen, it can be shared, so the computational overhead of averaging multiple snapshots of the same training is almost negligible (less than 10\% of the total forward time for averaging 3 predictions). Therefore, all our reported results are obtained by averaging 3 consecutive snapshots of the same training together (for example, results at 10M steps is the bagging of snapshots at 8M, 9M and 10M).

In summary, discrete actions can be compensated by increasing the number of actions, and averaging several discrete predictions.

\section{Experiments and Ablation Studies}
\label{results}

\subsection{Defining a Common Test Situation and a Metric for Comparison}

We first define a common set of scenarios and a metric to make fair comparison.
Indeed the CARLA challenge maps are not publicly available and the old CARLA benchmark is only available on a deprecated version of CARLA (0.8.X) on which rendering and physics differs from the version of CARLA used in the CARLA challenge (0.9.X). Moreover as aforementioned, this CARLA benchmark is a much simpler task than the CARLA challenge. 


\paragraph{Defining test scenarios} We choose the hardest environment in the available maps of CARLA. Town05 includes the biggest urban district, is mainly multi-lane and US style: the traffic lights are on the opposite side of the road and much harder to detect. We also randomly spawn pedestrians crossing the road ahead of our agent to verify our models brake on this situations. We additionally set changing weather to make the task as hard as possible. This way, even with a single town training, we have a challenging setup. The single town training is necessary to make all our experiments and ablations studies in a reasonable time. All those experiments were made with 20M iterations 
 on CARLA, with 3 actors (so 6.6M steps for each actor) and with a framerate of 10 FPS. Thus 20M steps is equivalent to around 20 days of simulated driving (as a comparison the most standard time \cite{mnih2015human, Dabney2018} used to train RL for Atari games is 200M frames corresponding to around 40 days and can go to more than 5 years of gametime \cite{R2D2, Horgan}). We define 10 scenarios of urban situations each one consisting in 10 consecutive intersections over the whole Town05 environment. We also define some scenarios on highway but those cases are much easier and thus less discriminative:  for example our best model goes off-road less than one time every 100km on highway situation. Highway scenarios are mostly used for evaluating the oscillations of our different agents.

\paragraph{Defining a metric to compare different model and ablation studies}
We test our models 10 times on each scenario varying the weather condition and resetting the position of all other agents. Contrary to the training phase, we only terminate episode when the agent goes off-road as this allows to keep track of the number of infractions encountered.
Our main metric is the average percentage of intersections successfully crossed (\textit{Inters.}, higher is better), for example 50\% completion corresponds to a mean of 5 intersections crossed in each scenario. We also keep track of the percentage of traffic lights passed without infraction (\textit{TL}, higher is better) and the percentage of pedestrians passed without collision (\textit{Ped.}, higher is better). Note that the last two are slightly less relevant, as a non-moving car will never run a red traffic light nor crash a pedestrian. That is why \textit{Inters.} is our main metric for comparison: \textit{TL} and \textit{Ped.} are used for more fine-grained comparison. We also introduce a measure for oscillations: the mean absolute rotation between the agent and the road along the episode (\textit{Osc.}, lower is better).

\subsection{Ablations Studies on the Supervised Phase}
\label{ablation_supervised}
In this section, we will detail our ablation studies concerning the supervised learning phase of affordances. The RL setup is exactly the same to ensure fair comparison.

\begin{figure*}[t]
    \centering
    \includegraphics[width=0.95\textwidth]{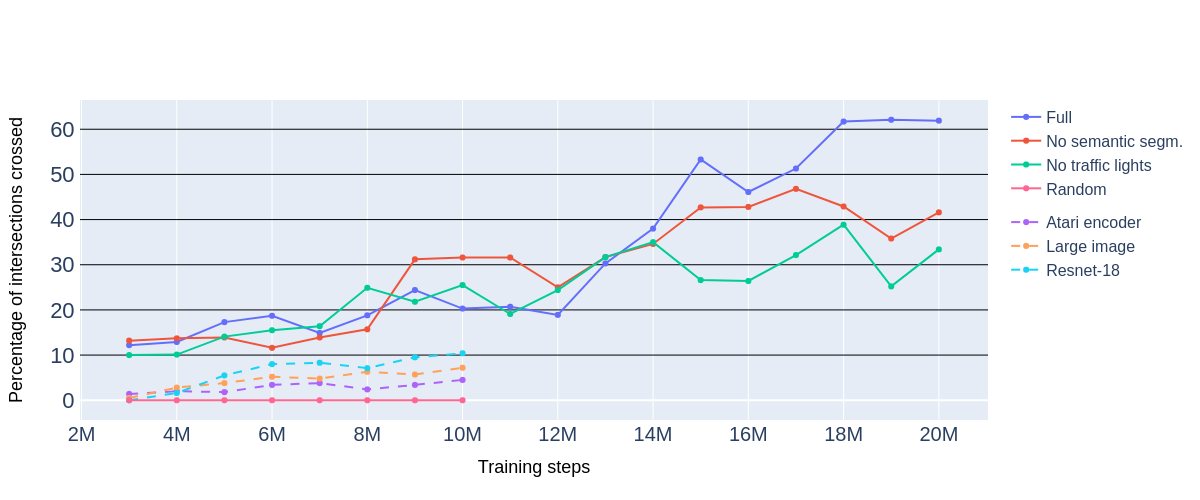}
    \caption{Evolution of agent performance with training steps and choice of the encoder behavior. The first group of encoders (solid lines) have frozen weights, the second group (dashed) are trained only by the RL signal (stopped earlier because the performance is clearly lower). Some experiments are averaged over multiple seeds (see Supplementary Materials for details on stability).}
    \label{fig:encoder_perfs}
\end{figure*}

First, some experiments are conducted without any supervised phase, i.e. training the whole network from scratch in the RL phase. Three different architectures are compared: the initial network from DQN with $84 \times 84$ images, a simple upgrade of the DQN network which takes $288 \times 288 \times 3$ images as input and finally our model with the Resnet-18 encoder.

Figure~\ref{fig:encoder_perfs} shows that without affordances learning, agents fail to learn and do not even succeed to pass one intersection in average (less than 10\% intersections crossed).
Moreover it is important to note that training the bigger image encoder (respectively the full resnet-18) took 50\% (resp. 200\%) more time than training with our implicit affordances scheme even considering the time used for the supervised phase. Consequently these experiments are stopped after 10M steps. These networks also require much more memory, because full images are stored in the replay memory. As expected, these experiments prove that training a large network using only RL signal is hard. 

\begin{table}[t]
    \centering
    \scriptsize
    \begin{tabular}{p{2.5cm}|*{3}{c}}
        \toprule
         Encoder used & Inters. & TL & Ped. \\
         \midrule
         Random & 0\% & NA & NA \\
         No TL state & 33.4\% & 80\% & 82\% \\
         No segmentation & 41.6\% & 96.5\% & 63\% \\
         All affordances & 61.9\% & 97.6\% & 76\% \\
         \bottomrule
    \end{tabular}
    \caption{Comparison of agent performance with regards to  encoder training loss (random weights, trained without  traffic light loss, without semantic segmentation loss, or with all affordance losses)}
    \label{tab:expe_on_supervised_phase}
\end{table}

The second stage of experiments concerned the Resnet-18 encoder training. First, as a sanity check, the encoder is frozen to random features. Then, either the traffic light state or the segmentation is removed from the loss in the supervised phase. These experiments show the interest of predicting the traffic light state and the semantic segmentation in our supervised training. The performance of the corresponding agents is illustrated in Figure~\ref{fig:encoder_perfs}.

Table~\ref{tab:expe_on_supervised_phase} shows that removing the traffic light state has a huge impact on the final performance. As expected the RL agent using an encoder trained without the traffic light loss is running more red traffic lights. It is interesting to note that this ratio is much better than a random choice (which would be 25\% of success for traffic light because traffic lights are green only 25\% of the time). This means that the agent still succeeds to detect some traffic light state signal in the features. We guess that as the semantic segmentation includes a traffic light class (but not the actual state of it) the features contain some information about traffic light state.
Removing the semantic segmentation loss from the encoder training also has an impact on final performance. As expected, performance on pedestrian collision is worse than any other training meaning the network has trouble to detect pedestrians and vehicles (this information is only contained in the semantic map).

\subsection{Ablations Studies on the RL Setup}
\label{ablation_RL}

For fair comparison, the same pre-trained encoder is used for all experiments, trained with all affordances mentioned in Section~\ref{frozen_encoder}. The encoder used here is the same one as the CARLA challenge, and has been trained on slightly more data and for more epochs than the encoders used for the previous ablation study.

Two experiments are conducted with different rewards to measure the impact of the reward shaping. In the first one (\textit{constant desired speed}), the desired speed is not adapted to the situation: the agent needs to understand only from termination signal to brake on red traffic lights and to avoid collisions. In the second experiment, the \textit{angle reward} component is removed to see the impact of this reward on oscillations. Two different settings for actions are also evaluated. First, the derivative of the steering angle is predicted instead of the current steering. Finally the steering angle discretization is studied, decreasing from 27 to 9 steering absolute values. Results are summarized in Table~\ref{tab:reward_shaping}.

\begin{table}[t]
    \centering
    \scriptsize
    \begin{tabular}{l|*{4}{c}}
        \toprule
         Input/output & Inters. & TL & Ped. & Osc. \\
         \midrule
         Constant desired speed & 50.3\% & 31\% & 42\% & 1.51\degree \\
         No angle reward & 64.7\% & 99\% & 77.7\% & 1.39\degree \\
         27 steering values (derivative) & 64.5\% & 98.7\% & 85.1\% & 1.64\degree \\
         9 steering values (absolute) & 74.4\% & 98.5\% & 84.6\% & 0.88\degree \\
         27 steering values (absolute) & 75.8\% & 98.3\% & 81.6\% & 0.84\degree \\
         \bottomrule
    \end{tabular}
    \caption{Performance comparison according to the steering angle discretization used and reward shaping}
    \label{tab:reward_shaping}
\end{table}

The most interesting result of these experiments is the one from \textit{Constant desired speed}. Indeed, the agent fails totally at braking for both cases of red traffic light or pedestrian crossing: its performance is much worse than any other agent. The agent trained with desired speed set to constant runs 70\% of traffic lights which is very close to a random choice. It also collides with 60\% of pedestrians. This experiment shows how important the speed reward component is to learn a braking behaviour. 

Surprisingly, we find that predicting derivative of steering results in more oscillations, even more than when removing the \textit{desired rotation} reward component. Finally, taking 9 or 27 different steering values does not have any significant impact and both of these agents reach the best performance with low oscillation.

\subsection{Generalization on Unseen Towns}

Finally, we conduct experiments on generalization, following the actual setting of the CARLA challenge. For this purpose, we train on 3 different towns at the same time (one with EU traffic light and the 2 others with US) and test on 2 unseen town (one EU and one US). We also test our best single town agent as a generalization baseline.

\begin{table}[t]
    \centering
    \scriptsize
    \begin{tabular}{p{2.1cm}|*{3}{p{0.8cm}}|*{3}{p{0.8cm}}}
        \toprule
         Training & \multicolumn{1}{c}{Unseen EU Town} & \multicolumn{1}{c}{Unseen US Town} \\
         Only Town05 & 2.4\% & 42.6\% \\
         Multi town & 58.4\% & 36.2\% \\
         \bottomrule
    \end{tabular}
    \caption{Generalization performance (\textit{Inters.} metric).}
    \label{tab:handling_discrete_actions}
\end{table}
\label{generalisation}

Results are presented in Table~\ref{tab:handling_discrete_actions}. We can see that performance on the unseen EU town is really poor for the agent trained only on a single US town, confirming the interest of training on both EU and US town at the same time. On the unseen US town, the performance is roughly similar for both trainings.
These experiments show that our method generalizes to unseen environments. 

\subsection{Comparison on CARLA Benchmark}

Very recently, Learning by Cheating (LBC) \cite{Chena} re-implemented on open-source the CARLA benchmark on the newest version of CARLA (0.9.6). 
With such limited time, we did not have time to change our training setup at time of submission regarding the weather condition, so only \textit{training weather} results are reported in Table~\ref{tab:bench_success} (test weather results can be found in the Supplementary).

\begin{table}[t]
 \centering
 \scriptsize
 \begin{tabulary}{\textwidth}{L*{5}{p{0.4cm}}|*{1}{p{0.6cm}}*{2}{p{0.4cm}}}
   \toprule
   & \multicolumn{5}{c}{CoRL2017 (train town)} & \multicolumn{3}{c}{NoCrash (train town)} \\
  Task & RL & CAL & CILRS & LBC & Ours & Task & LBC & Ours  \\
  \midrule
  Straight & 89 & 100 & 96 & 100 & 100 & Empty & 97 & 100 \\
  One turn & 34 & 97 & 92 & 100 & 100 & Regular & 93 & 96 \\
  Navigation & 14 & 92 & 95 & 100 & 100 & Dense & 71 & 70 \\
  Nav. dynamic & 7 & 83 & 92 & 100 & 100 & & & \\
  \bottomrule
 \end{tabulary} \\ \ \ \ \ \ \ \ \ \ \ \ \ \ \ \
  \begin{tabulary}{\textwidth}{L*{5}{p{0.4cm}}|*{1}{p{0.6cm}}*{2}{p{0.4cm}}}
   \toprule
   & \multicolumn{5}{c}{CoRL2017 (test town)} & \multicolumn{3}{c}{NoCrash (test town)} \\
  Task & RL & CAL & CILRS & LBC & Ours & Task & LBC & Ours  \\
  \midrule
  Straight & 74 & 93 & 96 & 100 & 100 & Empty & 100 & 99 \\
  One turn & 12 & 82 & 84 & 100 & 100 & Regular & 94 & 87 \\
  Navigation & 3 & 70 & 69 & 98 & 100 & Dense & 51 & 42 \\
  Nav. dynamic & 2 & 64 & 66 & 99 & 98 & & & \\
  \bottomrule
 \end{tabulary}
 \caption{Success rate comparison (in \% for each task and scenario, more is better) with baselines \cite{Dosovitskiy, Sauer, CILRS, Chena} on train weathers.}
 \label{tab:bench_success}
\end{table}

LBC \cite{Chena} which uses IL, is the only one outperforming our RL agent on the hardest task of \textit{CoRL2017} benchmark (ie. \textit{Nav. dynamic}). We also have similar results to the LBC baseline on the much harder \textit{NoCrash} benchmark. Note that we can only compare to LBC because other works have not been tested yet on the \textit{NoCrash} benchmark with pedestrians (only available in the open-source implementation of LBC \cite{Chena}). Finally, our work is outperforming the only other RL baseline \cite{Dosovitskiy} by a huge margin. This is also the first time a RL approach matches and even outperforms IL approaches on the CARLA benchmark. The inference code and the weights of our model can be found on open-source\footnote{https://github.com/valeoai/LearningByCheating}.

\section{Conclusion}

In this work, we present the first successful RL agent at end-to-end urban driving from vision including traffic light detection, using a value-based Rainbow-IQN-Apex training with an adapted reward and a large conditional network architecture. To solve this in a challenging autonomous driving context, we introduce \textit{implicit affordances}, which use a large encoder trained for tasks relevant to autonomous driving in a supervised setting. We validate our design choices with ablation studies, and showcased our performance by winning the track ``Camera Only" in the CARLA challenge.

In future work, it could be interesting to apply our \textit{implicit affordances} scheme for policy-based or actor-critic and to train our affordance encoder on real images in order to apply this method on a real car.

\section*{Acknowledgements}

The authors would like to thank Mustafa Shukor for his valuable time and his help on training some of our encoder.


{\small
\bibliographystyle{ieee_fullname}
\bibliography{carla_cvpr}
}

\clearpage

\begin{appendices}

\section{Supplementary materials: Implementation details}

In this section, we will detail the hyper-parameters and the architecture of both the Supervised and the Reinforcement Learning training.

\subsection{Supervised phase of affordances training: architecture and hyper-parameters}

Our encoder architecture is mainly based on Resnet-18 \cite{Resnet} with two main differences. First, we changed the first convolutional layer to take 12 channels as input (we stack 4 RGB frames). Secondly, we changed the kernel size of downsample convolutional layers from 1x1 to 2x2. Indeed as mentionned in the paper Enet \cite{Enet}, \textit{When downsampling,
the first 1x1 projection of the convolutional branch is performed with a stride of 2 in both dimensions,
which effectively discards 75\% of the input. Increasing the filter size to 2x2 allows to take the full
input into consideration, and thus improves the information flow and accuracy.}. We also removed the two last layers: the average pooling layer and the last fully connected. Finally, we added a last downsample layer taking 512x7x7 feature maps as input and outputting our RL state of size 512x4x4.

For the loss computation, we add a weight of 10 for the part of the loss around traffic light state detection, and 1 for all other losses.

\begin{table}[htbp]
\begin{center}
\caption{Supervised training hyperparameters}
\label{tab:hyperparameter_supervised}
\begin{tabular}{c|c}
Parameter & Value \\
\hline
Learning rate & $5.10^{-5}$, eps $3.10^{-4}$ (Adam) \\
Batchsize & 32 \\
Epochs & 20 \\
\end{tabular}
\end{center}
\end{table}

For the semantic decoder, each layer consists of an upsample layer with a nearest neighbor interpolation, then 2 convolutional layers with batchnorm. All the other losses are build with fully connected layers with one hidden layer of size 1024. See Table~\ref{tab:hyperparameter_supervised} for more details on other hyper-parameters used in the supervised phase.

To train our encoder, we used a dataset of around 1M frames with associated ground-truth label (e.g. semantic segmentation, traffic light state and distance). This dataset was collected mainly in 2 cities of the CARLA \cite{Dosovitskiy} simulator: Town05 (US) and Town02 (EU).

\subsection{Reinforcement Learning phase: architecture and hyper-parameters}

In all our RL trainings, we used our encoder trained on affordances learning as a frozen image encoder: the actual RL state is the 8162 features coming from this frozen encoder. We then give this state to one fully connected layer of size 8162x1024. Then from these 1024 features concatenated with the 4 previous speed and steering angle values, we use a gated network to handle different orders as presented in CIL \cite{Codevilla}. All the 6 heads have the same architecture but different weights, they are all made with 2 fully connected layers with one hidden layer of size 512.

\begin{table}[htbp]
\begin{center}
\caption{RL training hyperparameters for our \textit{Single Town} and \textit{Multi-Town} experiments: all parameters not mentioned come from the open-source implementation of Rainbow-IQN \cite{Toromanoff}.}
\label{tab:hyperparameter_RL}
\begin{tabular}{c|c}
Parameter & Single Town / Multi-Town \\
\hline
Learning rate & $5.10^{-5}$, eps $3.10^{-4}$ (Radam) \\
Batchsize & 32 \\
Memory capacity & 90 000 / 450 000 \\
Number actors & 3 / 9 \\
Number steps & 20M (23 days) / 50M (57 days) \\
Synchro. actors/learner & Yes / No \\
\end{tabular}
\end{center}
\end{table}

All hyperparameters used in our Rainbow-IQN training are the same as the one used in the open-source implementation \cite{Toromanoff} but for the replay memory size and for the optimiser. We use the really recent Radam \cite{Radam} optimiser as it is giving consistent improvement on standard supervised training. Some comparisons were made with the Adam optimiser but did not show any significant difference. For all our \textit{Single Town} experiments, we used Town05 (US) as environment. For our \textit{Multi-Town} training, we used Town02 (EU), Town04 (US) and Town05 (US). Table~\ref{tab:hyperparameter_RL} details the hyper-parameters used in our RL training.

\section{Experiments}

\subsection{Stability study}

One RL training of 20M steps was taking more than one week on a Nvidia 1080 Ti. That is why we did not have time nor computational resources to run an extensive study on the stability for all our experiments. Moreover evaluating our saved snapshot was also taking time, around 2 days to evaluate performance each million of steps as in Figure~7 of the main paper. Still, we performed multiple runs for 3 experiments presented in Table~1: \textit{No TL state}, \textit{No segmentation} and \textit{All Affordances}. We evaluated those seeds at 10M and at 20M steps and the results (mean and standard deviation) can be found in the following Table~\ref{tab:stability}.

\begin{table}[h]
    \centering
    \scriptsize
    \begin{tabular}{p{2cm}|*{2}{c}|*{2}{c}}
        \toprule
         & \multicolumn{2}{c}{10M steps} & \multicolumn{2}{c}{20M steps} \\
         Encoder used & Inters. & Nb seeds & Inters. & Nb seeds \\
         \midrule
         No TL state & 17.9\% $\pm$ 7.3 & 6 & 27\% $\pm$ 5.7 & 5 \\
         No segmentation & 27.7\% $\pm$ 9.3 & 5 & 41.7\% $\pm$ 0.1 & 2 \\
         All affordances & 24.9\% $\pm$ 8.2 & 6 & 64.4\% $\pm$ 2.5 & 2 \\
         \bottomrule
    \end{tabular}
    \caption{Mean and standard deviation of agents performance with regards to encoder training loss (trained without  traffic light loss, without semantic segmentation loss, or with all affordance losses)}
    \label{tab:stability}
\end{table}

Even if we just have few different runs, those experiments on stability support the fact that our training are roughly stable and our results are significant. At 20M steps the "best" seed of \textit{No TL state} perform worse than both seeds of \textit{No segmentation}. More importantly, both seeds of \textit{No segmentation} perform way worse than both seeds of \textit{All affordances}.

\subsection{Additional experiments}

We made one experiment, \textit{4 input one output}, to know the impact of predicting only one semantic segmentation instead of predicting 4 at the same time. Indeed, we stack 4 frames as our input and we thought it would give more information to learn from, if we train using all 4 semantic segmentations. We also tried to remove temporality in the input: taking only one frame as input and thus predicting only one semantic segmentation, \textit{One input one output}. Finally, we made an experiment, \textit{U-net Skip connection}, on which we used a standard U-net like architecture \cite{Unet} for the semantic prediction. Indeed we did not use skip connections in all our experiments to prevent the semantic information to flow in this skip connections. Our intuition was that the semantic information could not be present in our final RL state (the last features maps of 4x4) if using skip connections.

The results of this 3 experiments are described in Table~\ref{tab:additional_expe}. \\

\begin{table}[h]
    \centering
    \scriptsize
    \begin{tabular}{p{2.5cm}|*{3}{c}}
        \toprule
         Encoder used & Inters. & TL & Ped. \\
         \midrule
         One input one output & 29.6\% & 95\% & 85\% \\
         4 input one output & 64.3\% & 93.8\% & 70.7\% \\
         U-net Skip connection & 58.6\% & 95\% & 69.8\% \\
         All affordances & 64.4\% & 98.1\% & 76.2\% \\
         \bottomrule
    \end{tabular}
    \caption{Additional experiments to study impact of temporality both as input and as output of our Supervised phase. Also experiments with skip connection for the semantic prediction (U-net like skip connection \cite{Unet}).}
    \label{tab:additional_expe}
\end{table}

We can see from this results that using only one frame as input has a large impact on the final performance (going from 64\% intersections crossed with our standard scheme \textit{All Affordances} to 29\% when using only one image as input).
The impact of predicting only one semantic segmentation instead of 4 is marginal on our main metric (\textit{Inters.}) but we can see that the performance on traffic lights (\textit{TL}) and on pedestrians (\textit{Ped.}) are slightly lower.
Finally, the impact of using U-net like skip connections seems to be relatively small on the number of intersection crossed. However, there is still a difference with our normal system particularly on the pedestrians metric.

As a conclusion, those additional experiments confirmed our intuitions first about adding temporality both as input and output of our encoder and secondly to not use standard U-net skip connection is our semantic segmentation decoder to prevent semantic information to flow away from our final RL state. However, the impact of those intuitions are relatively small and we conducted only one seed which could not be representative enough.

\subsection{Description of our test scenario}

\begin{figure}[h]
    \centering
    \includegraphics[width=150px]{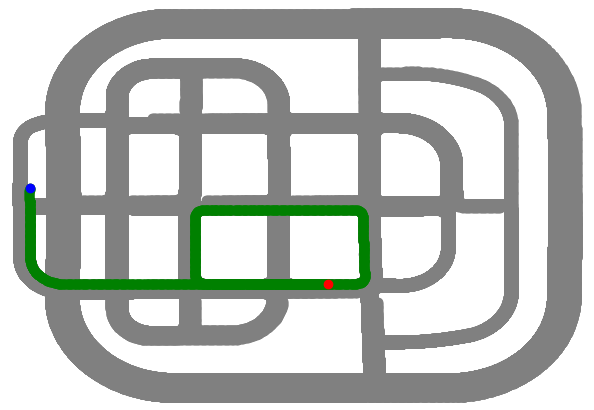} \\
    \caption{Sample of one of our scenario in Town05. The blue point is the starting point, the red is the destination.}
    \label{fig:scenario}
\end{figure}

Each of our scenario is defined by a starting waypoint and 10 orders one for each intersection to cross. An example of one of our 10 scenario can be found on Figure~\ref{fig:scenario}. We also spawn 50 vehicles in the whole Town05 while testing. Finally, we spawn randomly pedestrian ahead of the agent every 20/30 seconds. 

\subsection{Comparison on CARLA Benchmark: Implementation Details and Test Weathers Results}

\subsubsection{Test weathers results (train and test town)}

As mentioned in the main paper, we did not have time to re-implement our training setup for the really recently released \cite{Chena} implementation of the CARLA benchmark on the newer version of CARLA (0.9.6), particularly regarding the weather condition. At submission time, all our training were done under all possible weather conditions. That's why we reported our results only for training weathers in the main paper. We only had time to train our whole pipeline in the exact condition of the CARLA benchmark (i.e. only Town01 and train weathers for training and Town02 and test weathers for test) after acceptance. That's why we give our results for test weather only in the Supplementary Materials.

\begin{table}[ht]
 \centering
 \scriptsize
 \begin{tabulary}{\textwidth}{L*{5}{p{0.4cm}}|*{1}{p{0.6cm}}*{2}{p{0.4cm}}}
   \toprule
   & \multicolumn{5}{c}{CoRL2017 (train town)} & \multicolumn{3}{c}{NoCrash (train town)} \\
  Task & RL & CAL & CILRS & LBC & Ours & Task & LBC & Ours  \\
  \midrule
  Straight & 86 & 100 & 96 & 100 & 100 & Empty & 87 & 36 \\
  One turn & 16 & 96 & 96 & 96 & 100 & Regular & 87 & 34 \\
  Navigation & 2 & 90 & 96 & 100 & 100 & Dense & 63 & 26 \\
  Nav. dynamic & 2 & 82 & 96 & 96 & 100 & & & \\
  \bottomrule
 \end{tabulary} \\ \ \ \ \ \ \ \ \ \ \ \ \ \ \ \
  \begin{tabulary}{\textwidth}{L*{5}{p{0.4cm}}|*{1}{p{0.6cm}}*{2}{p{0.4cm}}}
   \toprule
   & \multicolumn{5}{c}{CoRL2017 (test town)} & \multicolumn{3}{c}{NoCrash (test town)} \\
  Task & RL & CAL & CILRS & LBC & Ours & Task & LBC & Ours  \\
  \midrule
  Straight & 68 & 94 & 96 & 100 & 100 & Empty & 70 & 24 \\
  One turn & 20 & 72 & 92 & 100 & 100 & Regular & 62 & 34 \\
  Navigation & 6 & 88 & 92 & 100 & 100 & Dense & 39 & 18 \\
  Nav. dynamic & 4 & 64 & 90 & 100 & 100 & & & \\
  \bottomrule
 \end{tabulary}
 \caption{Success rate comparison (in \% for each task and scenario, more is better) with baselines \cite{Dosovitskiy, Sauer, CILRS, Chena} on test weathers.}
 \label{tab:bench_success_supp}
\end{table}

We can see from Table~\ref{tab:bench_success_supp} that we are the only approach reaching a perfect score on all the tasks under test weathers.
However, we can see that our results on the \textit{NoCrash} benchmark fall far behind LBC \cite{Chena} baseline under test weathers (even if our results were similar under train weathers). We found that the test weathers on the \textit{NoCrash} benchmark are actually really different from the train weathers, particularly regarding sun reflection on the ground. We discovered that our frozen encoder trained only on Town01/train weathers was predicting sun reflection as "moving obstacles" and thus in this situation the RL agent is just braking for ever, acting like if a car was ahead. Most of our failure under test weathers on \textit{NoCrash} benchmark are in fact timeout because our agent is not moving anymore when he faces sun reflection on the ground. Handling diverse weather conditions is a known issue for perception algorithms and we think that improving our supervised performance (particularly the semantic segmentation) would probably manage this issue but this is left as future work.

\subsubsection{Implementation Details for the CARLA benchmark}

To train our new encoder in the exact condition of the CARLA benchmark, we used a new dataset of around 500K frames with associated ground-truth label (e.g. semantic segmentation, traffic light state and distance). This dataset was collected only in Town01 and under training weathers. Then we trained our RL agent with the \textit{implicit affordances} coming from this new encoder for around 40M steps using 9 actors with all actors on Town01 under training weathers. We used a slightly bigger field of view (from 90\degree \ to 100\degree) and we cropped the sky (from 288x288x3 images to 288x168x3) as the EU traffic lights are less high than the US traffic lights (the CARLA benchmark contains only EU traffic lights). Finally, we removed all the change lane orders because all towns in CARLA benchmark are single lane (the CARLA benchmark setup is actually simpler than the CARLA challenge for which this paper has been initially done).

\subsection{Training infrastructure}

The training of the agents was split over several computers and GPUs, containing in total:

\begin{itemize}
    \item 3 Nvidia Titan X and 1 Nvidia Titan V (training computer)
    \item 1 Nvidia 1080 Ti (local workstation)
    \item 2 Nvidia 1080 (local workstations)
    \item 3 Nvidia 2080 (training computer)
\end{itemize}

\end{appendices}

\end{document}